\documentclass{article}

\usepackage[preprint,nonatbib]{neurips_2022_m}

\usepackage[utf8]{inputenc} %
\usepackage[T1]{fontenc}    %
\usepackage{hyperref}       %
\usepackage{url}            %
\usepackage{booktabs}       %
\usepackage{amsfonts}       %
\usepackage{nicefrac}       %
\usepackage{microtype}      %
\usepackage{graphicx}
\usepackage{subfigure}
\usepackage{amssymb, amsmath}
\usepackage{stmaryrd}
\usepackage{wrapfig}
\usepackage{blindtext}
\usepackage{lipsum}
\usepackage[table]{xcolor}
\usepackage{longtable}
\usepackage{enumitem}
\usepackage{verbatimbox}
\usepackage{geometry}

\title{Novel View Synthesis for \\ High-fidelity Headshot Scenes}

\author{%
  Satoshi Tsutsui, Weijia Mao, Sijing Lin, Yunyi Zhu, Murong Ma, Mike Zheng Shou\thanks{Corresponding Author.} \\\\
  Show Lab, National University of Singapore
}

\begin{document}

\maketitle

\begin{abstract}
Rendering scenes with a high-quality human face from arbitrary viewpoints is a practical and useful technique for many real-world applications. Recently, Neural Radiance Fields (NeRF), a rendering technique that uses neural networks to approximate classical ray tracing, have been considered as one of the promising approaches for synthesizing novel views from a sparse set of images. We find that NeRF can render new views while maintaining geometric consistency, but it does not properly maintain skin details, such as moles and pores. These details are important particularly for faces because when we look at an image of a face, we are much more sensitive to details than when we look at other objects. On the other hand, 3D Morpable Models (3DMMs) based on traditional meshes and textures can perform well in terms of skin detail despite that it has less precise geometry and cannot cover the head and the entire scene with background. Based on these observations, we propose a method to use both NeRF and 3DMM to synthesize a high-fidelity novel view of a scene with a face. Our method learns a Generative Adversarial Network (GAN) to mix a NeRF-synthesized image and a 3DMM-rendered image and produces a photorealistic scene with a face preserving the skin details. Experiments with various real-world scenes demonstrate the effectiveness of our approach. The code will be available on \url{https://github.com/showlab/headshot} .
\end{abstract}

\section{Introduction}
Rendering scenes with a human face from arbitrary view is a useful technique for a variety of real-world applications including Augmented Reality (AR), Virtual Reality (VR), telepresence, games, etc. We are particularly interested in scenes with a face. Humans are extremely sensitive to the detail of faces than other objects, as we can discriminate faces from a very young age~\cite{goren1975visual}. In fact, a study shows that face dominates more than 10\% of visual stimuli for babies aged 1-11 months~\cite{jayaraman2015faces}, suggesting that we are naturally \textit{trained} to recognize faces in detail. Indeed, we can recognize even a slight difference if it is a face. Therefore, in this work, given training images of a scene with a face, we develop a technique that cannot only render scenes with a face from unseen views but can also keep the details of the face such as moles and pores. 

\begin{figure}[h!]
  \centering
    \includegraphics[width=\columnwidth]{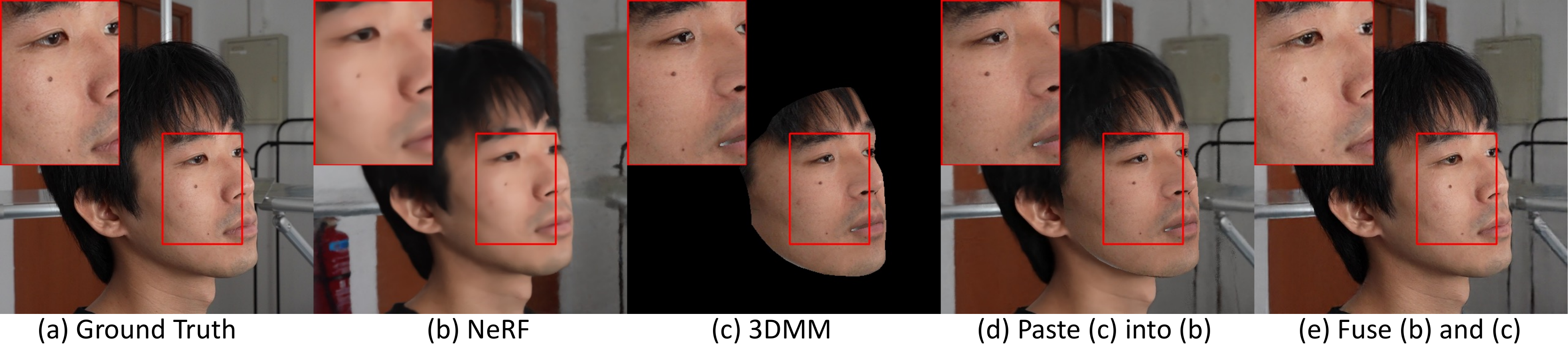}
    \caption{NeRF can render novel view like (b) with precise geometry and scene background but lacks skin details compared to the ground truth (a). 3DMM can render the frontal region of the face like (c) with skin details but with less accurate geometry and no background. Our idea is to make use of both NeRF and 3DMM to produce novel view of a scene with a face that maintains accurate geometry, background, and skin details. While naively pasting the 3DMM results into NeRF results produces a strange image like (d), we learn to fuse them and produce a novel view like (e).
    \label{fig:visstat}}%
\end{figure}

Human faces have been extensively studied in our community~\cite{egger20203d,zollhofer2018state}. For example, we can apply 3D Morpable Face Models (3DMMs)~\cite{DECA:Siggraph2021} to face images and obtain mesh and texture with skin details (see Figure~\ref{fig:3dmm}), which can be used to render the same face in different views. However, the rendered views are only for the frontal region of the face and do not include the entire head and hair, and there is no way to include the entire scene with backgrounds. Moreover, the view does not maintain precise geometric consistency and has slightly strange facial parts, such as eyes. An example of 3DMM rendering is shown in Figure~\ref{fig:visstat}(c). We see a strange eye and a slightly open mouth compared to the ground truth of Figure~\ref{fig:visstat}(a). \textbf{Observation 1}: \textit{3DMM-rendered faces maintain skin details but lack precise geometric consistency}. 

Beyond faces, recent progress in neural rendering achieved impressive performance by combining classical volume rendering techniques and the representation of 3D scenes approximated by neural networks, which are called Neural Radiance Fields (NeRF)~\cite{mildenhall2020nerf}. Given a set of training images of a complex scene, NeRF can render the scene in novel views while maintaining geometric consistency. However, we find that NeRFs often fail to capture details of the face, such as moles and pores. An example of NeRF rendering is shown in Figure~\ref{fig:visstat}(b). We see that the facial skins are blurry compared to the ground truth of Figure~\ref{fig:visstat}(a) . \textbf{Observation 2}: \textit{NeRF-rendered views of a scene with a face maintain the geometric consistency but lack skin details. }

\begin{figure}[t!]
  \centering
    \includegraphics[width=0.9\columnwidth]{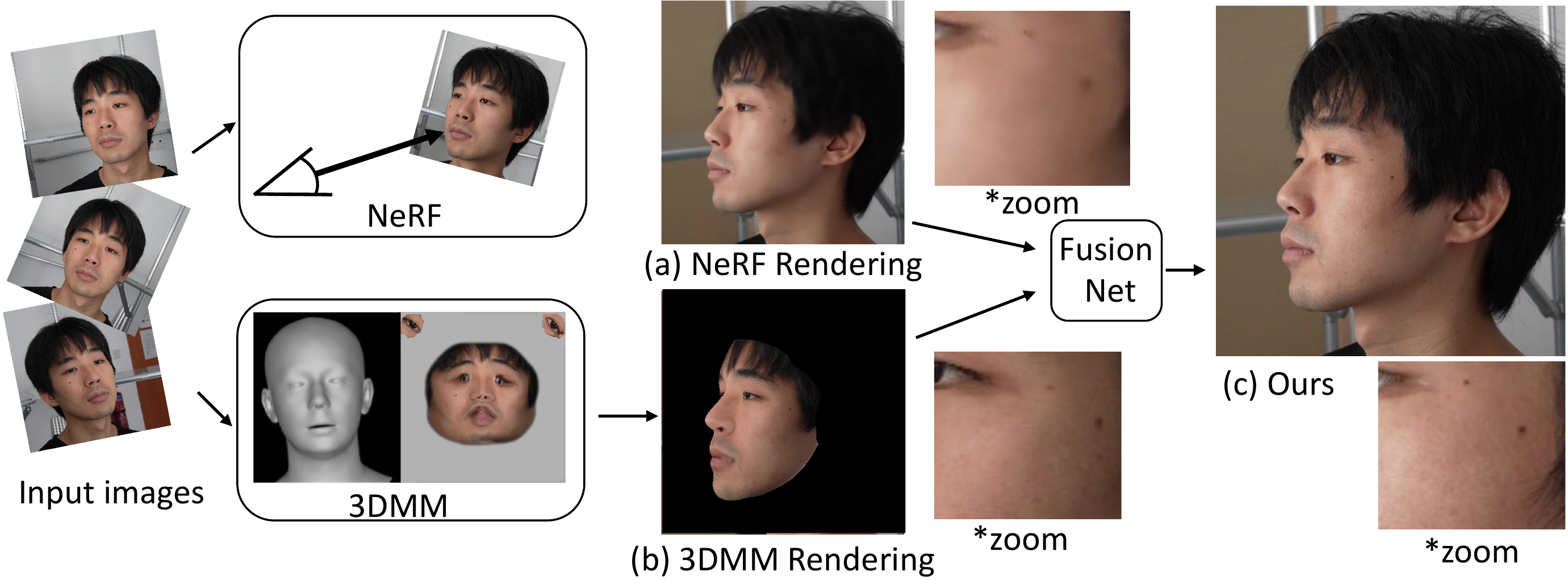}
    \caption{Given a sparse set of images from a scene with a non-rigid face, we synthesize high-fidelity novel views of the scene. As shown in Figure~\ref{fig:visstat}, NeRF can generate a view with geometric consistency and background but lacks skin details. 3DMM can only render the frontal region of the face with less accurate geometry but actually has more skin details. To take the advantages of the both models, we use FusionNet to produce an image like (c) with accurate geometry, background, and skin details.   \label{fig:method}}
\end{figure}

\textbf{Proposed work.} Based on observations 1 and 2, we would like to render a face with geometric consistency, background, and skin details using NeRF and 3DMM. Given a short video clip of a scene with a face, our method effectively applies NeRF and 3DMMs to synthesize novel views while keeping both the geometric consistency and the skin details of the face (see Figure~\ref{fig:method}). Our idea is to generate novel views with geometric consistency with NeRF, and then recover the details of the skin with 3DMM. Naively mixing the 3DMM face into the NeRF rendered image produces a face image like in Figure~\ref{fig:visstat}(d), which has some artifacts of ``pasteing''. We do the mixing job much better without artifacts. To do so, we propose learning to mix them using Convolutional Neural Networks (CNNs) with Generative Adversarial Network (GANs), which we call Fusion Net. The Fusion Net, which is trained adversarially~\cite{goodfellow2014generative}, takes a novel view from NeRF and a 3DMM face rendering and produces the novel view with geometric consistency and skin details as in Figure~\ref{fig:visstat}(e). We experimentally demonstrate the performance of our method using various real-world scenes with a face. Our data include faces with various expressions, speaking faces, indoor scenes, outdoor scenes, and complex backgrounds. 

\textbf{Contributions}. 1) We propose a method to synthesize novel high-fidelity views for scenes with a face, while keeping geometric consistency and skin details.  2) We effectively fuse 3DMM (a prior-based 3D face model) and NeRF (a neural 3D model without prior) benefiting from the strengths of the two methods. 3) Experimentally demonstrate the effectiveness of our approach using real-world scenes with various conditions.

\section{Related Work}
Our work is related to novel view synthesis and 3D modeling of human face. We do not cover full-body modeling because it usually cannot capture the details of the face.

\paragraph{Novel View Synthesis and Neural Rendering}
Novel view synthesis, as a long-standing task in the field of computer vision, is a task of synthesizing images of a scene from a free view point given a sparse set of views.  Traditional explicit 3D representations include point clouds~\cite{agarwal2011building, liu2010pointcloud}, voxels \cite{girdhar2016learning, yan2016perspective}, meshes \cite{Matsuyama2004realtime, tung2009complete} and surfles \cite{pfister2000surfels, carceroni2002multi}. Recently, representing 3D data implicitly via multi-layer perceptrons (MLPs) turned out to be effective and is called NeRF~\cite{mildenhall2020nerf}. It was originally designed for static scenes, but has recently been extended for dynamic scenes \cite{pumarola2021dnerf, park2021nerfies, tretschk2021nonrigid,du2021neural}. Our work uses frames from a monocular video, so we want to use models that can handle dynamic scenes. Although any NeRF model for dynamic scenes can be used, we use a simple one called D-NeRF \cite{pumarola2021dnerf}. While NeRF and its dynamic variants are for generic scenes, NerFace~\cite{GafniNerface21} is a model particularly for the secne with a face, and feeding facial expression embedding into the MLPs. We compare with NerFace as a baseline. 

\paragraph{Human Face Modeling}
Modeling 3D faces has been one of the core problems in computer vision and computer graphics for decades. A pioneering work is the classic 3D Morphable Face Model (3DMM) ~\cite{blanz1999morphable} (see survey~\cite{egger20203d} for details and history). It reconstructs a 3D face from multi-view images and then applies a linear Principal Component Analysis (PCA) to represent a new face instance with a linear combination of bases. Constructing accurate face models often requires a specialized capturing studio with multi-view geometry~\cite{ghosh2011multiview,li2021tofu}. Recent work~\cite{bagautdinov2018modeling,tewari2018self,tran2018nonlinear,yenamandra2021i3dmm,zhuang2021mofanerf,ma2021pixel,DECA:Siggraph2021} applies deep learning methods including neural rendering to infer accurate 3D face models (see a survey~\cite{zollhofer2018state}). They are different from our work in that they are not for novel view synthesis of the scene with a face (e.g., they do not care the scene background). Meanwhile, we want a model that can estimate the mesh and texture from a face image. Although our method can adapt any work that can estimate mesh and textures, we use DECA~\cite{DECA:Siggraph2021} due to its effective performance and the high-quality implementations available.

\section{Method}
Our method takes a sparse set of views from a scene with a non-rigid face and renders novel views with geometric consistency and details of the skin (Figure~\ref{fig:method}). We used a monocular camera to capture a short (less than a few minutes) video clip of a dynamic scene with a face by manually moving the camera around the subject. We use COLMAP~\cite{schoenberger2016sfm} to obtain the intrinsic and extrinsic camera parameters. We use a neural rendering model to synthesize a novel view with geometric consistency but without skin details (Sec~\ref{sec:nerf};Figure~\ref{fig:method}a). We also use a 3D morphable face model to render the frontal region of the face with skin details but without precise geometric consistency (Sec~\ref{sec:3dmm};Figure~\ref{fig:method}b). To synthesize novel views with geometric consistency and skin details (Figure~\ref{fig:method}c), we fuse the results of the two models using Fusion Networks (Sec~\ref{sec:fusionnet}). 

\subsection{Rendering with Dynamic Neural Radiance Field}\label{sec:nerf}
Given a set of images with camera parameters, we use Neural Radiance Fields (NeRF)~\cite{mildenhall2020nerf} to represent the scene and render images from an arbitrary points of view using a volumetric rendering technique. The original NeRF is designed for static scenes, but the scenes we use are dynamic, so we use its dynamic variant of D-NeRF~\cite{pumarola2021dnerf}.

\paragraph{Scene Representation} We use neural networks to approximate the volume opacity $\sigma$ and the color $\mathbf{c} = (r,g,b)$ of a 3D location $\mathbf{x} = (x,y,z)$ viewed from a direction $\mathbf{d}$ at a time $t$. Given $(\mathbf{c}, \mathbf{d}, t)$, the networks $F$ estimate $(\sigma, \mathbf{c})$, so $F:(\mathbf{x}, \mathbf{d}, t) \mapsto (\sigma, \mathbf{c})$. For the design choice of $F$, we follow D-NeRF~\cite{pumarola2021dnerf}, which demonstrated that using two networks is better than having a single network for approximating a dynamic radiance field. The network $F$ contains the canonical network $F_{\theta}$ and the deformation network $F_{\tau}$. The first network $F_{\theta}$ represents the canonical space that does not depend on time. $F_{\theta}$ takes the inputs of $(\mathbf{x},\mathbf{d})$ and outputs $(\sigma,\mathbf{c})$. To incorporate the time information, we also use the second network $F_{\tau}$ that maps a location at a time into the location in the canonical space. The mapping is represented as a displacement $\Delta \mathbf{x}$ from a location $\mathbf{x}$ at a time $t$ and its corresponding position in the canonical space as $\mathbf{x} + \Delta\mathbf{x}$. $F_{\tau}$ takes the inputs of ($\mathbf{x}$, $t$) and outputs $\Delta\mathbf{x}$. $F_{\theta}$ and $F_{\tau}$ are multilayer perceptrons (MLPs). Rather than 
directly feeding $\mathbf{x}, \mathbf{d}$ and $t$ into MLPs, we use Positional Encoding~\cite{vaswani2017attention} $\gamma(p) = \left[\sin(2^l\pi p),\cos(2^l\pi p)\right]_{l=0}^{L-1}$, which is known to be better than using direct values~\cite{mildenhall2020nerf}. 

\paragraph{Volumetric Rendering} Using the dynamic radiance field defined above, we render an image corresponding to camera parameters (intrinsics and extrinsics) using a ray tracing technique. To render a pixel $\hat{C}_{t}(\mathbf{r})$ of the image at the time of $t$, we march a camera ray $\mathbf{r}(s) = \mathbf{o} + s\mathbf{d}$ with the optical center $\mathbf{o}$ and the direction $\mathbf{d}$ computed from the camera parameters.

\begin{eqnarray}
    {C}_{t}(\mathbf{r}) = 
        \int_{s_n}^{s_f} T(s) \;
        \sigma( \mathbf{r}(s), t) \;
        \mathbf{c}( \mathbf{r}(s), \mathbf{d}, t) \;
        ds ,
    \\
    \;\mathrm{where}\; 
    T(s) = 
        \exp \left(
            -\int_{s_n}^{s}
            \sigma( \mathbf{r}(h), t)  
            dh
            \right), 
\end{eqnarray}
$T(s)$ is the accumulated opacity of the ray from $s_n$ to $s$ (that is, the probability that the ray traverses from $s_n$ to $s$ without being blocked), $\sigma( \mathbf{r}(s), t)$ and $\mathbf{c}( \mathbf{r}(s), \mathbf{d}, t)$ are the outputs of $ F(\mathbf{r}(s), \mathbf{d}, t)$, and the scene is bounded by the near-far of $[s_n,s_f]$. The above equations are approximated by sampling discrete sets $\{s_i\}_{i=1}^{N}$ between $s_n$ and $s_f$.

\begin{eqnarray}
    \hat{C}_{t}(\mathbf{r})  = 
        \sum_{i=1}^{N} T_i\left( 1-\exp(-\sigma_i\delta_i)\mathbf{c_i}\right),
        \;\mathrm{where}\;
        T_i = \exp\left( -\sum_{j=1}^{i-1}\sigma_j\delta_j \right), 
\end{eqnarray}
$(\sigma_i, \mathbf{c}_i) = F(\mathbf{r}(s_i), \mathbf{d}, t)$, and $\delta_i =  s_{i+1} - s_{i}$ is the distance between the neighboring samples. 

\paragraph{Loss Function} By sampling a batch of rays $\mathcal{R}$ from the training images with corresponding camera parameters, we minimize the L2 loss between the pixel value of the ground truth $C_{\mathrm{gt}}(\mathbf{r})$ and the predicted value $\hat{C}_{t}(\mathbf{r})$.

\begin{eqnarray}
\min_{F} \sum_{\mathbf{r} \in \mathcal{R}}  \left\Vert \hat{C}_{t}(\mathbf{r}) - C_{\mathrm{gt}}(\mathbf{r})  \right\Vert^2_{2}
\end{eqnarray}

\subsection{Mesh and Texture from 3D Morphable Face Model (3DMM)}\label{sec:3dmm}
\begin{figure}
  \centering
    \includegraphics[width=\columnwidth]{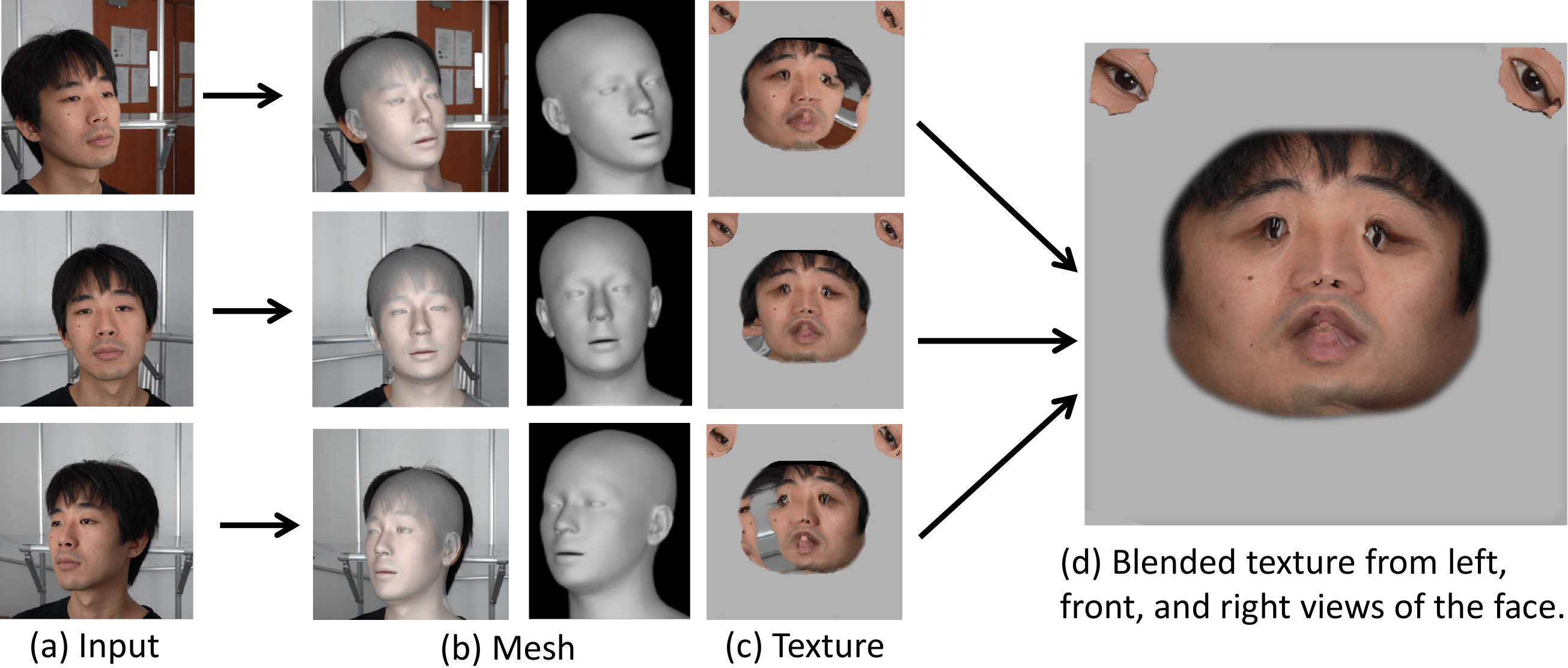}
    \caption{3DMM fitting and texture stitching\label{fig:3dmm}. Because a single view cannot cover the all of the texture UV space, we use left, middle, and right views, and combine them to obtain a complete texture.}
\end{figure}

We fit a 3DMM model into the training images to get the 3D shape and the texture.  We use a recent model of DECA~\cite{DECA:Siggraph2021} that uses deep neural networks to estimate the parameters of a latest 3DMM model called FLAME~\cite{FLAME:SiggraphAsia2017}. Given an image of a face, DECA gives the estimated mesh and the projected locations of its vertices in the image. We use piecewise affine transformation~\cite{affine_transform} to map the projected vertex locations in the image into the corresponding locations defined in the texture map, and obtain the texture of the human face.
DECA runs on a single image, but we cannot obtain the texture map covering the entire UV space from a single image, because a single view cannot cover all of the vertices of the frontal face. Hence we use the three images corresponding to the right, front, and left faces, and then blend the resulting texture maps into a complete texture map by exploiting overlapping parts. This texture stitching method is similar to Bao et al.\cite{bao2021high}. The process of the 3DMM fitting and texture stitching is visualized in Figure~\ref{fig:3dmm}.

\subsection{Detail Recovery with Fusion Net}\label{sec:fusionnet}
While NeRF (Sec.~\ref{sec:nerf}) can render the images from novel views with precise geometry (shown in Figure \ref{fig:method}a), they lack details of the skin. Meanwhile, by fitting the 3DMM model~\ref{sec:3dmm} into the NeRF outputs, we can obtain recover the skin details (shown in Figure \ref{fig:method}b). However, the resulting face is only for the frontal region without the hair, head, and background, and is not precisely aligned with the geometry of the face. For example, the locations of the eyes are a little bit shifted to the right than the actual location. Indeed, the 3DMM results maintain many more skin details, such as freckles, pores, and moles. Therefore, we use neural networks to fuse the NeRF results (Figure \ref{fig:method}a) and 3DMM results (Figure \ref{fig:method}b) and obtain the enhanced results that have precise geometry and skin details (Figure \ref{fig:method}c). We name this neural network as Fusion Net. 

\paragraph{Fusion Net} $F_{\phi}$ takes the input of NeRF-synthesized image $I_N$ (e.g., Figure \ref{fig:method}a) and its corresponding 3DMM image $I_M$ (e.g., Figure \ref{fig:method}b), and outputs the fused images (e.g., Figure \ref{fig:method}c). Our design choice of $F_{\phi}$ is inspired by the image-to-image translation framework~\cite{pix2pix2017} with Generative Adversarial Network (GANs)~\cite{goodfellow2014generative}, and particularly Pix2PixHD~\cite{wang2018pix2pixHD}. $F_{\phi}$ is a convolutional neural network (CNN) composed of an encoder network with convolutional layers, convolutional blocks with residual connections~\cite{he2016deep}, and a decoder network with transposed convolution layers. For inputting the two images into the $F_{\phi}$, we concatenate them in the channel dimension and feed the six-channel image $\mathbf{x} = \mathrm{concat}([I_N,I_M])$. %

\paragraph{Loss Function} We optimize $F_{\phi}$ using the L2 loss function and the adversarial loss function $\mathcal{L}_{adv}$ using a CNN-based discriminator $D$ (a binary classifier) and the ground truth image $I_{gt}$ as follows.

\begin{equation}
    \min_{F_{\phi}} \left( \max_D \mathcal{L}_{adv}(F_{\phi},D) + \lambda_2 \left\Vert F_{\phi}(\mathbf{x}) - I_{gt} \right\Vert^2_{2} \right), 
    \label{eq:fusionloss}
\end{equation}
\begin{equation}
\;\mathrm{where}\; \mathcal{L}_{adv}(F_{\phi},D) = \mathop{{}\mathbb{E}}[\log\left(D(I_{gt}, \mathbf{x} )\right)] + \mathop{{}\mathbb{E}}[\log\left(1-D(F_{\phi}(I_{gt},\mathbf{x}))\right)],
\end{equation}
and $\lambda_2$ is a hyperparameter to decide the weight of the L2 loss.

\section{Experiments}

\begin{table}[tb!]
  \caption{We capture seven real-world scenes with a face under various conditions of face and background. }
  \label{tbl:dataset}
  \centering
  \begin{tabular}{cllll}
    \toprule
    Name     & Face & Mouth & Background & Place \\
    \midrule
    Scene 1 & Light Expression Change &  Closed & Simple textured wall & Outdoor\\ %
    Scene 2 & Light Expression Change &  Closed  & Complex leaves & Outdoor\\ %
    Scene 3 & Light Expression Change &  Closed  & Distant buildings and trees & Outdoor\\ %
    Scene 4 & Middle Expression Change &  Closed &  Wall and objects at sides & Indoor\\ %
    Scene 5 & Extreme Expression Change&  Open & Locker and dolls & Indoor\\
    Scene 6 & Speaking &  Open & Cluttered shelf with toys etc & Indoor\\ %
    Scene 7 & Speaking &  Open & Building and complex leaves  & Outdoor\\ %
    \bottomrule
  \end{tabular}
\end{table}

\begin{figure}[tb!]
  \centering
    \includegraphics[width=\columnwidth]{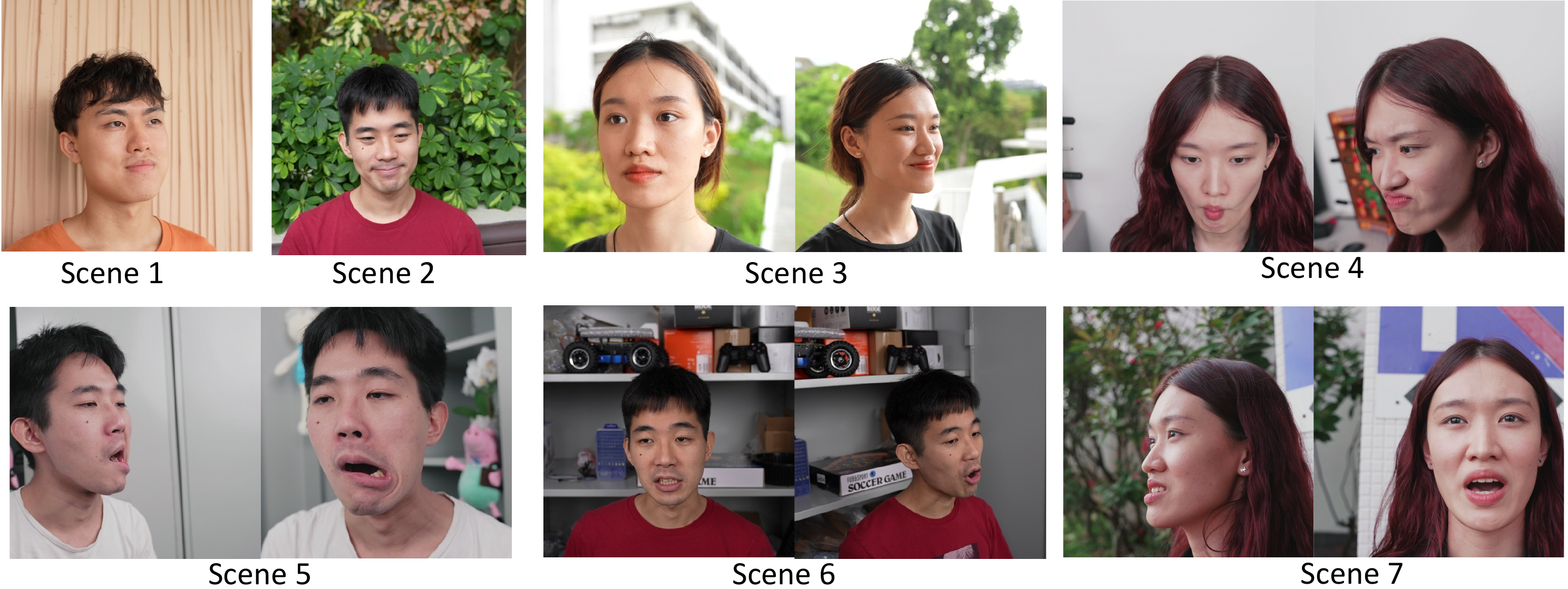}
    \caption{Our dataset has seven scenes with various conditions of face and background. See Table~\ref{tbl:dataset} for details.\label{fig:dataset}}
\end{figure}

\begin{figure}[htb!]
  \centering
    \includegraphics[width=0.92\columnwidth]{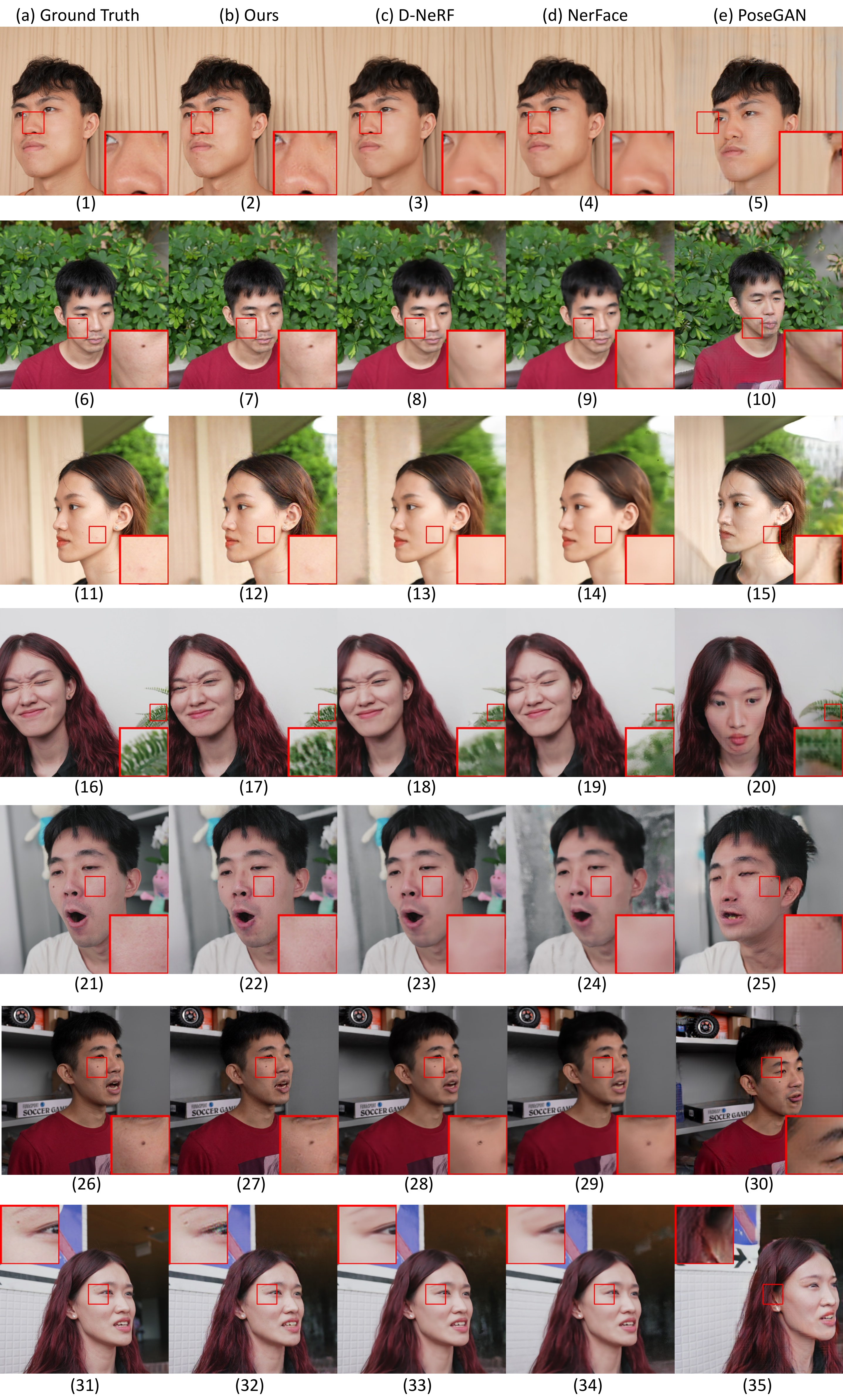}
    \caption{Testing results of the seven scenes described in Table~\ref{tbl:dataset} and Figure~\ref{fig:dataset}. The smaller red boxes of each row are located in the exactly same image coordinates, and the larger ones are zoomed boxes. Our method keeps more details than the others. See Sec.~\ref{sec:case} for the discussion. \label{fig:vis1}}
\end{figure}

\subsection{Datasets}
We use a camera to capture seven real-world scenes with a face under various conditions of face and background. We manually move the camera around a face (see sample frames in Figure~\ref{fig:dataset}). A point of our work is to care for the details of the face such as moles or pores, so we use Sony Alpha 7S III, a very high quality mirrorless camera. We also explored GoPro and Insta360 at the initial stage, but found that their quality is not as high as that of Sony. We take a short (a few minutes at maximum) video clip in 4K resolution with 60FPS with hardware stabilization enabled. We later down-sampled the temporal resolution because the views between temporally adjacent frames are often almost identical. We describe more details of this process in the supplementary material. We do a center crop of the captured images and resize them to $512 \times 512$. We found that 512 is the minimum resolution for which we can care for the skin details of the face, while saving time to wait for training of the neural networks compared to the original resolution. Faces in our dataset have light-weight expression changes (e.g., smile), middle expression changes (e.g., eyebrow or cheeks in), or extreme expression changes (e.g., mouth largely opened). For the middle and extreme expressions, we used those defined in previous work on face modeling~\cite{COMA:ECCV18}. The background can be cluttered with complex objects (e.g., leaves) or very distant. Our data include indoor and outdoor scenarios. We summarize the characteristics of the seven scenes in Table~\ref{tbl:dataset}. We refer the supplementary material for more details about the dataset.

\subsection{Implementation Details}\label{sec:impdetails}
The canonical network $F_{\theta}$ and the deformation network $F_{\tau}$ are the eight layered MLPs with 256 neurons and ReLU non-linearity, following the D-NeRF~\cite{pumarola2021dnerf}. We use the scene at $t=0$ as the canonical space, so $F_{\tau}(\mathbf{x},0)=\mathbf{0}$. We use the Positional Encoding with $L=10, 4,$ and $ 4$ for $\mathbf{x}, \mathbf{d}$ and $t$, respectively. We use the Adam~\cite{kingma2014adam} optimizer with the learning rate $0.0005$ for $800000$ iterations with the batch size of $500$ rays. The training takes around two days with GeForce RTX 2080 Ti. The Fusion Network $F_{\phi}$ has five convolutional layers in the encoder, four residual blocks, and five transposed convolution layers in the decoder. The discriminator $D$ has five convolutional layers.  We set $\lambda_2 = 1$ in equation~\ref{eq:fusionloss}. We use the Adam optimizer with the initial learning rate $0.0002$ for $60000$ iterations with the batch size of $8$. The training takes around half a day with GeForce RTX 2080 Ti. More implementation details are provided in the supplementary material.

\subsection{Baselines}
We compare our method with the following baselines.
\textbf{D-NeRF}~\cite{pumarola2021dnerf} is a basis for our model and uses neural fields and volumetric rendering to synthesize novel views conditioned on time. \textbf{NerFace}~\cite{GafniNerface21} is a baseline of neural fields and volumetric rendering that are specifically designed for faces. Instead of time, it takes the 3DMM facial expression code, where we use DeCA~\cite{DECA:Siggraph2021}. \textbf{PoseGAN} is a baseline for synthesizing novel views using GANs conditioned on camera poses. Since our method uses GANs to recover the details of the face, it is natural to include the baseline of GANs only. However, this baseline was not easy to realize. We originally customized FastGAN~\cite{liu2021towards} (a GAN designed for a small number of images; our training images are less than a thousand) by replacing the input random noise with learnable pose embeddings, but the training collapsed. The non-collapsing approach we found was to train FastGAN conditioned on random noise and then learn a linear mapping from camera pose to GAN latent space similar to the GAN Inversion~\cite{xia2021gan}. This way enables the model to output a deterministic image for each camera pose, which is required to act as a novel view synthesis. Lastly, we prepare \textbf{Ours + Blur} that applies a $3\times3$ Gaussian filter to our results in order to show that some evaluation metrics are biased to blurry images  (see Sec.~\ref{sec:results-metrics}). 

\subsection{Qualitative Results}\label{sec:case}
We sample a single moment per scene and visualize them in Figure~\ref{fig:vis1}. PoseGAN is obviously worse than the others and fails to render the accurate geometry, which can be confirmed by the fact that red bounding boxes (pointing to the same locations for each row) of PoseGAN correspond to the different part of the face compared to the others. Hence, in the remainder of this paragraph, we only discuss the results of Ours, D-NeRF, and NerFace. In general, we can visually confirm that our method preserves more skin details than the others, as we can see in Figure~\ref{fig:vis1}-(2), (7), (12), (22), (27) and (32). NerFace results are mostly similar to D-NeRF in that the skin details are blurry. Not only skin details, our approach often renders more details of the background. For example, the background of Figure~\ref{fig:vis1}-(11), (12), and (13) tell us that our method has a clearer background than D-NeRF and is closer to the ground truth background. Because our Fusion Net takes the D-NeRF results as an input, we can say that Fusion Net sometimes recovers missing details for non-facial regions. However, unlike faces, Fusion Net does not take the texture of the background, so it may introduce the details that does not exist. For example, by comparing Figure~\ref{fig:vis1}-(16) and (17), we can see that the leaf patterns are changed.

\subsection{Quantitative Results}\label{sec:results-metrics}
We follow previous work~\cite{mildenhall2020nerf} to report Peak Signal-to-Noise Ratio (PSNR- higher is better), Structural Similarity Index Measure (SSIM-  higher is better), and Learned Perceptual Image Patch Similarity (LPIPS - lower is better), as shown in Table~\ref{tbl:results}. We report the average of these metrics per scene and across the seven scenes. For PSNR and SSIM, we do not observe a consistent trend to judge the best method, except that PoseGAN is the worst. We believe that PSNR and SSIM are not the best metric as we care about the details of the skin, such as pores. D-NeRF and Nerface results are often blurry, whereas ours are not. Hence, we tried a $3\times3$ Gaussian filter to our results (Our + Blur) and discovered consistent improvements in PSNR and SSIM. Our + Blur has the highest PSNR and SSIM (29.36 and 0.8520) on average across the seven scenes, indicating that these metrics are not appropriate here and that we may want to perform a human evaluation to access the quality (see the next section).  Meanwhile, LPIPS is the state-of-the-art metric based on deep learning that is claimed to correlate well with human judgment~\cite{zhang2018perceptual}. Our method achieves the highest LPIPS in all scenes. Moreover, LPIPS scores consistently rank as Ours > D-NeRF >  NerFace > PoseGAN. This indicates the effectiveness of our method. 

\setlength{\tabcolsep}{3.5pt}
\begin{table}[tb!]
\caption{Qualitative comparison of the seven scenes with a face. Ours are consistently better in terms of LPIPS. See Sec.~\ref{sec:results-metrics} for the discussion.  }\label{tbl:results}
\resizebox{\linewidth}{!}
{   
    \begin{tabular}{@{}lllllllllllll@{}}
    \toprule
    \multicolumn{1}{c}{} & \multicolumn{3}{c}{Scene 1}      & \multicolumn{3}{c}{Scene 2} & \multicolumn{3}{c}{Scene 3} & \multicolumn{3}{c}{Scene 4} \\
    Method     & PSNR$\uparrow$ & SSIM$\uparrow$ & LPIPS$\downarrow$ & PSNR$\uparrow$ & SSIM$\uparrow$ & LPIPS$\downarrow$  & PSNR$\uparrow$ & SSIM$\uparrow$ & LPIPS$\downarrow$  & PSNR$\uparrow$ & SSIM$\uparrow$ & LPIPS$\downarrow$   \\ 
    \cmidrule(r){1-1}\cmidrule(lr){2-4}\cmidrule(lr){5-7}\cmidrule(lr){8-10}\cmidrule(lr){11-13}
     PoseGAN     &  15.49 &  0.4966 &  0.3733 &  13.49 &  0.2486 &  0.4410 &  12.30 &  0.4954 &  0.5119 &  12.07 &  0.5133 &  0.5258 \\
	NerFace    &  30.33 &  0.8422 &  0.1328 &  \textbf{25.78} &  \textbf{0.7697} &  0.1849 &  20.99 &  0.7365 &  0.3202 &  28.75 &  0.8412 &  0.2176 \\
	D-NeRF     &  30.35 &  \textbf{0.8625} &  0.0913 &  23.22 &  0.7021 &  0.1798 &  26.20 &  0.8313 &  0.2158 &  \textbf{30.26} &  \textbf{0.8720} &  0.1498 \\
    \arrayrulecolor{black!20}
    \midrule
    \arrayrulecolor{black}
	Ours       &  30.10 &  0.8268 &  \textbf{0.0669} &  24.09 &  0.7165 &  \textbf{0.0880} &  28.67 &  0.8497 &  \textbf{0.0760} &  29.47 &  0.8230 &  \textbf{0.1017} \\
    Ours + Blur &  \textbf{30.78} &  0.8531 &  0.0953 &  24.46 &  0.7411 &  0.1172 &  \textbf{29.24} &  \textbf{0.8713} &  0.1022 &  30.02 &  0.8545 &  0.1225 \\
    \arrayrulecolor{black!20}
    \midrule
    \arrayrulecolor{black}
	Ours w/o 3DMM &  30.04 &  0.8210 &  0.0894 &  24.08 &  0.7145 &  0.0898 &  28.82 &  0.8482 &  0.0805 &  29.40 &  0.8126 &  0.1166 \\
	Ours w/o FusionNet &  27.03 &  0.8190 &  0.1031 &  24.20 &  0.7526 &  0.1233 &  24.20 &  0.8005 &  0.2209 &  26.66 &  0.8270 &  0.1574 \\

	\midrule
	\\
	\midrule
	
	\multicolumn{1}{c}{} & \multicolumn{3}{c}{Scene 5}      & \multicolumn{3}{c}{Scene 6} & \multicolumn{3}{c}{Scene 7} & \multicolumn{3}{c}{Mean}    \\
    Method     & PSNR$\uparrow$ & SSIM$\uparrow$ & LPIPS$\downarrow$ & PSNR$\uparrow$ & SSIM$\uparrow$ & LPIPS$\downarrow$  & PSNR$\uparrow$ & SSIM$\uparrow$ & LPIPS$\downarrow$  & PSNR$\uparrow$ & SSIM$\uparrow$ & LPIPS$\downarrow$   \\ 
    \cmidrule(r){1-1}\cmidrule(lr){2-4}\cmidrule(lr){5-7}\cmidrule(lr){8-10}\cmidrule(lr){11-13}
    PoseGAN     &  14.18 &  0.5874 &  0.5318 &  14.34 &  0.4801 &  0.4951 &  13.05 &  0.4485 &  0.4949 &  13.56 &  0.4671 &  0.4820 \\
	NerFace    &  24.46 &  0.8133 &  0.3230 &  29.81 &  0.8963 &  0.1118 &  27.53 &  0.8151 &  0.2831 &  26.81 &  0.8163 &  0.2248 \\
	D-NeRF      &  30.25 &  \textbf{0.9028} &  0.1541 &  \textbf{32.52} &  \textbf{0.9319} &  0.0736 &  \textbf{28.63} &  \textbf{0.8485} &  0.1640 &  28.78 &  0.8502 &  0.1469 \\
    \arrayrulecolor{black!20}
	\midrule
	\arrayrulecolor{black}
	Ours      &  29.82 &  0.8627 &  \textbf{0.1060} &  31.73 &  0.9008 &  \textbf{0.0416} &  28.03 &  0.8046 &  \textbf{0.0977} &  28.84 &  0.8263 &  \textbf{0.0826} \\
    Ours + Blur  &  \textbf{30.32} &  0.8899 &  0.1273 &  32.21 &  0.9188 &  0.0723 &  28.49 &  0.8354 &  0.1266 &  \textbf{29.36} &  \textbf{0.8520} &  0.1091 \\
    \arrayrulecolor{black!20}
    \midrule
    \arrayrulecolor{black}
	Ours w/o 3DMM &  29.74 &  0.8562 &  0.1190 &  31.68 &  0.8993 &  0.0453 &  28.03 &  0.7999 &  0.1160 &  28.83 &  0.8217 &  0.0938 \\
	Ours w/o FusionNet &  26.13 &  0.8481 &  0.1591 &  29.68 &  0.9065 &  0.0695 &  25.61 &  0.8140 &  0.1641 &  26.22 &  0.8240 &  0.1425 \\
    \bottomrule
    \end{tabular}
}
\end{table}
\begin{table}[t!]
  \caption{Mean rating with 95\% confidence interval from the user study.}
  \label{tbl:userstudy}
  \centering
  \begin{tabular}{cccc}
    \toprule
    Ours     & D-NeRF & NerFace & PoseGAN  \\
    \midrule
    $ 3.331 \pm 0.134 $ &$ 3.075 \pm 0.094 $ &$ 2.569 \pm 0.125 $ &$ 1.006 \pm 0.012 $ \\
    \bottomrule
  \end{tabular}
\end{table} 

\subsection{User Study}\label{sec:userstudy}
To further assess the effectiveness of our approach, we perform a user study to rate the quality of the four methods: our method and the three baselines of D-NeRF, NerFace, and PoseGAN. We collected 14 subjects, showed the results of the four methods and asked them to rate from 1 (the worst), 2 (second worst), 3 (the second best), 4 (the best). We showed 10 random moments per subject, while taking at least one moment from each of the seven scenes. In total, we have $14 \times 10 = 140$ rating samples per method. We compute the average rating and 95\% confidence interval for each method, as shown in Table~\ref{tbl:userstudy}. Our method achieved the rating of $ 3.331 \pm 0.134 $, which is significantly higher than the other three baselines. We also confirm that the trend of the rate is the same as that of LPIPS in that Ours > D-NeRF >  NerFace > PoseGAN.

\subsection{Ablations}\label{sec:ablations}
We investigate the two ablations of our model. \textbf{Ours w/o 3DMM} is a model to use Fusion Net~\ref{sec:fusionnet} only with D-NeRF-reduced images without 3DMM~\ref{sec:3dmm} images. \textbf{Ours w/o FusionNet} is a model to overlay (or simply \textit{paste}) the 3DMM rendered images on the corresponding D-NeRF rendered images. We show the PSNR, SSIM, and LPIPS in Table~\ref{tbl:results}. We look at LPIPS because our user study confirmed that it correlates with human judgments. The trend of LPIPS is consistent with Ours > (Ours w/o 3DMM) > (Ours w/o FusionNet) for all scenes. This indicates that both 3DMM and FusionNet contribute our method. 

\section{Conclusion}
We proposed a novel view synthesis method for a scene with a face. Our method takes advantage of NeRF and 3DMM. NeRF can render the entire scene with geometric consistency but lacks skin details of the face. Meanwhile, 3DMM can only render a frontal part of the face with less geometric consistency, but keeps skin details from the texture. Our method fuses them, benefits from both, and renders the entire scene with a face with skin details. We realize it by learning to mix the NeRF and 3DMM results through adversarial training. Our experiments, including a user study, demonstrate the effectiveness of our approach. A limitation of our work is that it is a novel view synthesis so cannot render the unseen expressions of human face. Our method can change the expression of the subject by changing the time, but the expression has to be observed at a time.

\clearpage
\bibliographystyle{plain}
\bibliography{references}

\appendix

\section{Demo Video}
Please check \url{https://github.com/showlab/headshot} .

\section{Implementation Details}
The canonical network $F_{\theta}$ and the deformation network $F_{\tau}$ are based on D-NeRF~\cite{pumarola2021dnerf}. Our implementation is based on their code\footnote{\url{https://github.com/albertpumarola/D-NeRF}}. Following their model definitions\footnote{\url{https://github.com/albertpumarola/D-NeRF/blob/f16319df497105b71ac151d2c2ddd4de36a1493f/run_dnerf_helpers.py\#L68-L126}}
\footnote{\url{https://github.com/albertpumarola/D-NeRF/blob/f16319df497105b71ac151d2c2ddd4de36a1493f/run_dnerf_helpers.py\#L142-L210}} in pytorch, the specific network architectures of $F_{\theta}$ and $F_{\tau}$ are the below. 

\begin{verbnobox}[\fontsize{7pt}{7pt}\selectfont]
DirectTemporalNeRF(
  (_occ): NeRFOriginal(
    (pts_linears): ModuleList(
      (0): Linear(in_features=63, out_features=256, bias=True)
      (1): Linear(in_features=256, out_features=256, bias=True)
      (2): Linear(in_features=256, out_features=256, bias=True)
      (3): Linear(in_features=256, out_features=256, bias=True)
      (4): Linear(in_features=256, out_features=256, bias=True)
      (5): Linear(in_features=319, out_features=256, bias=True)
      (6): Linear(in_features=256, out_features=256, bias=True)
      (7): Linear(in_features=256, out_features=256, bias=True)
    )
    (views_linears): ModuleList(
      (0): Linear(in_features=283, out_features=128, bias=True)
    )
    (feature_linear): Linear(in_features=256, out_features=256, bias=True)
    (alpha_linear): Linear(in_features=256, out_features=1, bias=True)
    (rgb_linear): Linear(in_features=128, out_features=3, bias=True)
  )
  (_time): ModuleList(
    (0): Linear(in_features=84, out_features=256, bias=True)
    (1): Linear(in_features=256, out_features=256, bias=True)
    (2): Linear(in_features=256, out_features=256, bias=True)
    (3): Linear(in_features=256, out_features=256, bias=True)
    (4): Linear(in_features=256, out_features=256, bias=True)
    (5): Linear(in_features=319, out_features=256, bias=True)
    (6): Linear(in_features=256, out_features=256, bias=True)
    (7): Linear(in_features=256, out_features=256, bias=True)
  )
  (_time_out): Linear(in_features=256, out_features=3, bias=True)
)
\end{verbnobox}

For $F_{\theta}$ and $F_{\tau}$, we use the Adam~\cite{kingma2014adam} optimizer with the learning rate $0.0005$ for $800000$ iterations with the batch size of $500$ rays. The learning rate is exponentially decayed to $0.00005$. The training takes around two days with GeForce RTX 2080 Ti.

The Fusion Network $F_{\phi}$ and the discriminator $D$ are based on Pix2PixHD~\cite{wang2018pix2pixHD}, following its public implementation\footnote{\url{https://github.com/NoelShin/Pix2PixHD}}. Following their model definitions\footnote{\url{https://github.com/NoelShin/Pix2PixHD/blob/9b157c24f2c89368d3188cdadfeeb514aaa563ad/networks.py}}, the specific network architectures of $F_{\phi}$ and $D$ are the below. 

\begin{verbnobox}[\fontsize{7pt}{7pt}\selectfont]
FusionNet(
  (model): Sequential(
    (0): ZeroPad2d((3, 3, 3, 3))
    (1): Conv2d(6, 32, kernel_size=(7, 7), stride=(1, 1))
    (2): InstanceNorm2d(32, eps=1e-05, momentum=0.1, affine=False, track_running_stats=False)
    (3): ReLU(inplace=True)
    (4): Conv2d(32, 64, kernel_size=(3, 3), stride=(2, 2), padding=(1, 1))
    (5): InstanceNorm2d(64, eps=1e-05, momentum=0.1, affine=False, track_running_stats=False)
    (6): ReLU(inplace=True)
    (7): Conv2d(64, 128, kernel_size=(3, 3), stride=(2, 2), padding=(1, 1))
    (8): InstanceNorm2d(128, eps=1e-05, momentum=0.1, affine=False, track_running_stats=False)
    (9): ReLU(inplace=True)
    (10): Conv2d(128, 256, kernel_size=(3, 3), stride=(2, 2), padding=(1, 1))
    (11): InstanceNorm2d(256, eps=1e-05, momentum=0.1, affine=False, track_running_stats=False)
    (12): ReLU(inplace=True)
    (13): Conv2d(256, 512, kernel_size=(3, 3), stride=(2, 2), padding=(1, 1))
    (14): InstanceNorm2d(512, eps=1e-05, momentum=0.1, affine=False, track_running_stats=False)
    (15): ReLU(inplace=True)
    (16): ResidualBlock(
      (block): Sequential(
        (0): ZeroPad2d((1, 1, 1, 1))
        (1): Conv2d(512, 512, kernel_size=(3, 3), stride=(1, 1))
        (2): InstanceNorm2d(512, eps=1e-05, momentum=0.1, affine=False, track_running_stats=False)
        (3): ReLU(inplace=True)
        (4): ZeroPad2d((1, 1, 1, 1))
        (5): Conv2d(512, 512, kernel_size=(3, 3), stride=(1, 1))
        (6): InstanceNorm2d(512, eps=1e-05, momentum=0.1, affine=False, track_running_stats=False)
      )
    )
    (17): ResidualBlock(
      (block): Sequential(
        (0): ZeroPad2d((1, 1, 1, 1))
        (1): Conv2d(512, 512, kernel_size=(3, 3), stride=(1, 1))
        (2): InstanceNorm2d(512, eps=1e-05, momentum=0.1, affine=False, track_running_stats=False)
        (3): ReLU(inplace=True)
        (4): ZeroPad2d((1, 1, 1, 1))
        (5): Conv2d(512, 512, kernel_size=(3, 3), stride=(1, 1))
        (6): InstanceNorm2d(512, eps=1e-05, momentum=0.1, affine=False, track_running_stats=False)
      )
    )
    (18): ResidualBlock(
      (block): Sequential(
        (0): ZeroPad2d((1, 1, 1, 1))
        (1): Conv2d(512, 512, kernel_size=(3, 3), stride=(1, 1))
        (2): InstanceNorm2d(512, eps=1e-05, momentum=0.1, affine=False, track_running_stats=False)
        (3): ReLU(inplace=True)
        (4): ZeroPad2d((1, 1, 1, 1))
        (5): Conv2d(512, 512, kernel_size=(3, 3), stride=(1, 1))
        (6): InstanceNorm2d(512, eps=1e-05, momentum=0.1, affine=False, track_running_stats=False)
      )
    )
    (19): ResidualBlock(
      (block): Sequential(
        (0): ZeroPad2d((1, 1, 1, 1))
        (1): Conv2d(512, 512, kernel_size=(3, 3), stride=(1, 1))
        (2): InstanceNorm2d(512, eps=1e-05, momentum=0.1, affine=False, track_running_stats=False)
        (3): ReLU(inplace=True)
        (4): ZeroPad2d((1, 1, 1, 1))
        (5): Conv2d(512, 512, kernel_size=(3, 3), stride=(1, 1))
        (6): InstanceNorm2d(512, eps=1e-05, momentum=0.1, affine=False, track_running_stats=False)
      )
    )
    (20): ConvTranspose2d(512, 256, kernel_size=(3, 3), stride=(2, 2), padding=(1, 1), output_padding=(1, 1))
    (21): InstanceNorm2d(256, eps=1e-05, momentum=0.1, affine=False, track_running_stats=False)
    (22): ReLU(inplace=True)
    (23): ConvTranspose2d(256, 128, kernel_size=(3, 3), stride=(2, 2), padding=(1, 1), output_padding=(1, 1))
    (24): InstanceNorm2d(128, eps=1e-05, momentum=0.1, affine=False, track_running_stats=False)
    (25): ReLU(inplace=True)
    (26): ConvTranspose2d(128, 64, kernel_size=(3, 3), stride=(2, 2), padding=(1, 1), output_padding=(1, 1))
    (27): InstanceNorm2d(64, eps=1e-05, momentum=0.1, affine=False, track_running_stats=False)
    (28): ReLU(inplace=True)
    (29): ConvTranspose2d(64, 32, kernel_size=(3, 3), stride=(2, 2), padding=(1, 1), output_padding=(1, 1))
    (30): InstanceNorm2d(32, eps=1e-05, momentum=0.1, affine=False, track_running_stats=False)
    (31): ReLU(inplace=True)
    (32): ZeroPad2d((3, 3, 3, 3))
    (33): Conv2d(32, 3, kernel_size=(7, 7), stride=(1, 1))
    (34): Tanh()
  )
)
Discriminator(
  (Scale_0): PatchDiscriminator(
    (block_0): Sequential(
      (0): Conv2d(9, 64, kernel_size=(4, 4), stride=(2, 2), padding=(1, 1))
      (1): LeakyReLU(negative_slope=0.2, inplace=True)
    )
    (block_1): Sequential(
      (0): Conv2d(64, 128, kernel_size=(4, 4), stride=(2, 2), padding=(1, 1))
      (1): InstanceNorm2d(128, eps=1e-05, momentum=0.1, affine=False, track_running_stats=False)
      (2): LeakyReLU(negative_slope=0.2, inplace=True)
    )
    (block_2): Sequential(
      (0): Conv2d(128, 256, kernel_size=(4, 4), stride=(2, 2), padding=(1, 1))
      (1): InstanceNorm2d(256, eps=1e-05, momentum=0.1, affine=False, track_running_stats=False)
      (2): LeakyReLU(negative_slope=0.2, inplace=True)
    )
    (block_3): Sequential(
      (0): Conv2d(256, 512, kernel_size=(4, 4), stride=(1, 1), padding=(1, 1))
      (1): InstanceNorm2d(512, eps=1e-05, momentum=0.1, affine=False, track_running_stats=False)
      (2): LeakyReLU(negative_slope=0.2, inplace=True)
    )
    (block_4): Sequential(
      (0): Conv2d(512, 1, kernel_size=(4, 4), stride=(1, 1), padding=(1, 1))
    )
  )
  (Scale_1): PatchDiscriminator(
    (block_0): Sequential(
      (0): Conv2d(9, 64, kernel_size=(4, 4), stride=(2, 2), padding=(1, 1))
      (1): LeakyReLU(negative_slope=0.2, inplace=True)
    )
    (block_1): Sequential(
      (0): Conv2d(64, 128, kernel_size=(4, 4), stride=(2, 2), padding=(1, 1))
      (1): InstanceNorm2d(128, eps=1e-05, momentum=0.1, affine=False, track_running_stats=False)
      (2): LeakyReLU(negative_slope=0.2, inplace=True)
    )
    (block_2): Sequential(
      (0): Conv2d(128, 256, kernel_size=(4, 4), stride=(2, 2), padding=(1, 1))
      (1): InstanceNorm2d(256, eps=1e-05, momentum=0.1, affine=False, track_running_stats=False)
      (2): LeakyReLU(negative_slope=0.2, inplace=True)
    )
    (block_3): Sequential(
      (0): Conv2d(256, 512, kernel_size=(4, 4), stride=(1, 1), padding=(1, 1))
      (1): InstanceNorm2d(512, eps=1e-05, momentum=0.1, affine=False, track_running_stats=False)
      (2): LeakyReLU(negative_slope=0.2, inplace=True)
    )
    (block_4): Sequential(
      (0): Conv2d(512, 1, kernel_size=(4, 4), stride=(1, 1), padding=(1, 1))
    )
  )
)
\end{verbnobox}

For $F_{\phi}$, we use the Adam optimizer with the initial learning rate $0.0002$ for $60000$ iterations with the batch size of $8$. The learning rate is exponentially decayed to $0.0000002$. The training takes around half a day with GeForce RTX 2080 Ti. 

\section{Dataset Details}
We take a short (a few minutes at maximum) video clip of each scene with 60FPS, but later down-sample the temporal resolution because the views between temporally adjacent frames are often almost identical. The down-sampled rates are different for each clip, ranging from 2 to 16. The rate is decided by manually checking that the adjacent frames are not nearly identical. The final number of frames used for the experiments are 62, 115, 502, 82, 82, 161, and 89 for the Scene 1, 2, 3, 4, 5, 6, and 7 respectively. We use the 16\% of the frames for testing. 

\end{document}